\documentclass{article}
\usepackage{amsmath,epsfig}
\usepackage{amsfonts,amssymb,bbm}
\usepackage{color}
\usepackage[preprint]{spconfa4}

\copyrightnotice{978-1-7281-1331-9/20/\$31.00 ©2020 IEEE}

\let\OLDthebibliography\thebibliography
\renewcommand\thebibliography[1]{
  \OLDthebibliography{#1}
  \setlength{\parskip}{0pt}
  \setlength{\itemsep}{0pt plus 0.3ex}
}

\begin{document}\sloppy
\topmargin=0mm

\def\x{{\mathbf x}}
\def\L{{\cal L}}

\title{Constrained R-CNN: A general image manipulation detection model}
%
\name{Chao Yang$^{\ast}$, Huizhou Li$^{\ast}$, Fangting Lin$^{\ast}$, Bin Jiang$^{\ast}$, Hao Zhao$^{\dagger}$}
\address{$^{\ast}$College of Computer Science and Electronic Engineering, Hunan University \\\{yangchaoedu, lihz, linfangting, jiangbin\}@hnu.edu.cn\\$^{\dagger}$Alipay (Hangzhou) Information \& Technology Co., Ltd.  zhaohao.zh@antfin.com}

\maketitle

\begin{abstract}
Recently, deep learning-based models have exhibited remarkable performance for image manipulation detection. However, most of them suffer from poor universality of handcrafted or predetermined features. Meanwhile, they only focus on manipulation localization and overlook manipulation classification. To address these issues, we propose a coarse-to-fine architecture named Constrained R-CNN for complete and accurate image forensics. First, the learnable manipulation feature extractor learns a unified feature representation directly from data. Second, the attention region proposal network effectively discriminates manipulated regions for the next manipulation classification and coarse localization. Then, the skip structure fuses low-level and high-level information to refine the global manipulation features. Finally, the coarse localization information guides the model to further learn the finer local features and segment out the tampered region. Experimental results show that our model achieves state-of-the-art performance. Especially, the $F_1$ score is increased by 28.4\%, 73.2\%, 13.3\% on the NIST16, COVERAGE, and Columbia dataset.
\end{abstract}
\section{Introduction}
\label{sec:intro}

\begin{figure}
	\centering
	\includegraphics[width=1\linewidth]{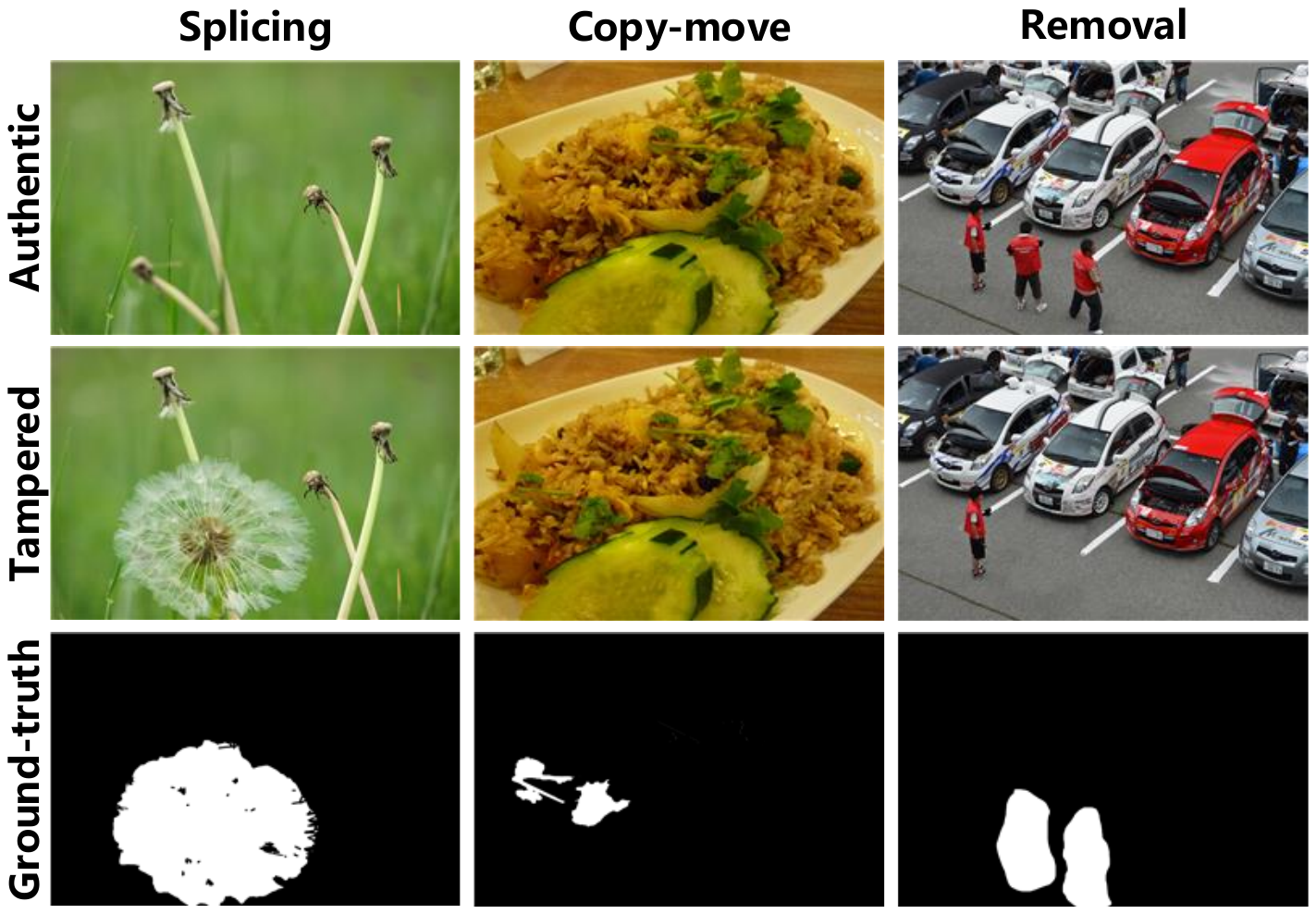}
	\vspace{-2em}
	\caption{Example of tampered images with different content manipulation techniques. From left to right are different manipulation techniques as splicing, copy-move, and removal.}
	\label{fig:fig1}
	\vspace{-1em}
\end{figure}

Image manipulation has been considered as a potential threat, negatively affecting many aspects of our life, such as fake news, bogus certificate, malicious rumors, etc. As shown in Fig. \ref{fig:fig1}, content manipulation is the most harmful forgery type, because it changes the image content subtly through various techniques containing splicing, copy-move, and removal. To shield this threat, image manipulation detection has received extensive attention, which aims to identify the authenticity of the content in an image without any prior knowledge.

Early works mainly utilize handcrafted or predetermined features such as frequency domain characteristics \cite{4285009} and color filter array (CFA) pattern \cite{1511009} to detect manipulated images. However, these methods cannot apply to real forensics because the features they employ are always specific-defined for one type of manipulation technique. Recently, deep learning methods are applied to this task, which improves the generalizability of models \cite{DBLP:conf/eccv/HuhLOE18,DBLP:conf/cvpr/BunkBMNFMCRP17,DBLP:conf/cvpr/ZhouHMD18,DBLP:journals/tip/BappySNMR19,DBLP:conf/cvpr/0001AN19}. These models first extract predetermined features, and then these features are individually \cite{DBLP:conf/cvpr/BunkBMNFMCRP17} or together \cite{DBLP:conf/cvpr/ZhouHMD18,DBLP:journals/tip/BappySNMR19,DBLP:conf/cvpr/0001AN19} fed into the deep learning model to classify whether a patch or pixel is tampered or not. Nevertheless, these methods only mitigate the issue of insufficient generalization ability, they are still fundamentally limited by handcrafted or predetermined features.

Additionally, we argue that image manipulation detection contains two goals as “\textbf{How}” (manipulation techniques classification) and “\textbf{Where}” (manipulation localization), which are equally important in this task. Such as the copy-move image in Fig. \ref{fig:fig1}, if only the mask is provided, we will probably mistake it as a splicing image. Therefore, if the model only focuses on one goal, the results of forensics will not be convincing. However, most of recent methods \cite{DBLP:journals/jvcir/SalloumRK18,DBLP:conf/iccv/BappyRBNM17,DBLP:conf/cvpr/0001AN19} only focus on “\textbf{Where}” but overlook “\textbf{How}”. In \cite{DBLP:conf/cvpr/ZhouHMD18}, although the model considers two goals, it uses bounding box to coarsely localize manipulation rather than more precise pixel-wise prediction.
\begin{figure*}
	\centering
	\includegraphics[width=0.95\linewidth]{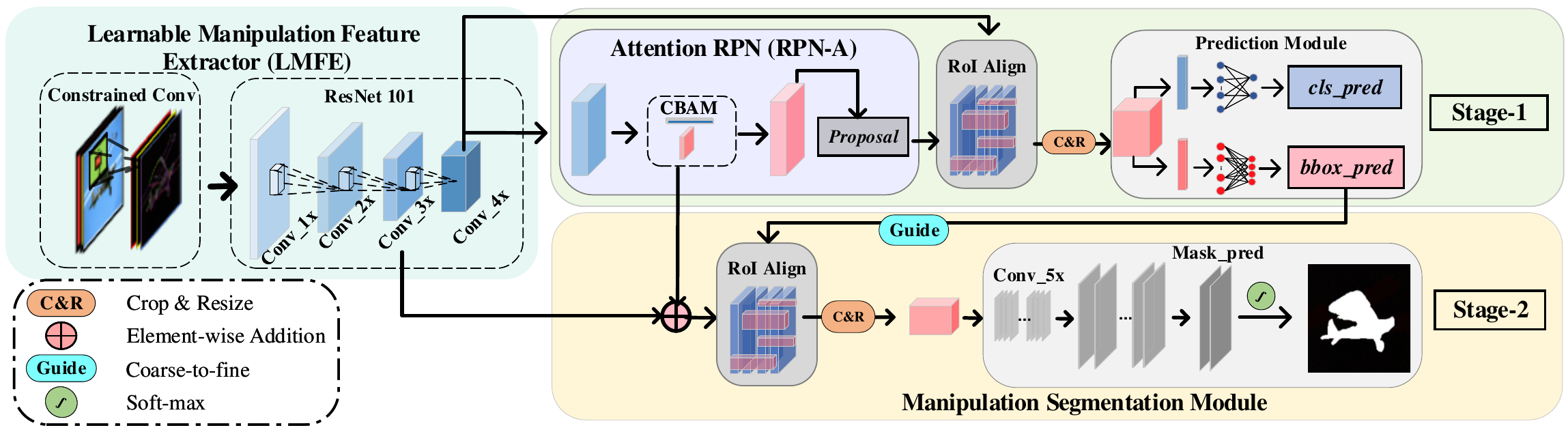}
	\vspace{-1em}
	\caption{Overview of the proposed architecture for image manipulation detection.}
	\label{fig:fig2}
	\vspace{-1em}
\end{figure*} 

According to the above analysis, there exists two major issues in existing works: 1) handcrafted or predetermined features limit the generalization ability of models, 2) overlook the integrity of forensics task. This motivates us to design a general solution that applies to detect various content manipulation, which achieves manipulation techniques classification and tampered region segmentation simultaneously.

In this paper, we propose Constrained R-CNN, a general, end-to-end and effective architecture for image manipulation detection. For the first issue, we design a learnable manipulation feature extractor (LMFE) based on the constrained convolution layer \cite{DBLP:journals/tifs/BayarS18} to create a unified feature representation of various content manipulation directly from data. For the second issue, we design a two-stages architecture to simulate the coarse-to-fine forensic process in the real world. More specifically, we optimize the Mask R-CNN \cite{DBLP:conf/iccv/HeGDG17} on task level, which consists of two stages: 1) Stage-1, we design an attention region proposal network (RPN-A) to identify the manipulated regions for latter manipulation classification and coarse localization. 2) Stage-2,  the skip structure fuses the low-level and high-level information to enhance the global feature representation. As prior knowledge, the bounding box information from Stage-1 guide Stage-2 to focus on the local feature in bounding box for final tampered region segmentation.

Our main contributions are as follows: 1) A coarse-to-fine architecture named Constrained R-CNN that achieves manipulation techniques classification and manipulated region segmentation simultaneously. 2) A single-stream learnable manipulation feature extractor, which creates a unified feature representation of various content manipulation techniques directly from data. 3) An attention regional proposal network (RPN-A), which discriminates manipulated regions effectively. Experiments on four benchmark datasets show that our model achieves state-of-the-art performance.

\section{Proposed Method}

\subsection{Overview}
In real-world forensics, expert identifies an image following a coarse-to-fine process. In the coarse stage, the expert roughly observes the image for approximate manipulation localization. In the fine stage, a more detailed analysis of specific regions, looking for more clues to segment tampered regions. 

Inspired by this process, we propose a two-stage architecture named Constrained R-CNN to simulate the coarse-to-fine process. As shown in Fig. \ref{fig:fig2}, Constrained R-CNN mainly consists of the learnable manipulation feature extractor (LMFE), the coarse manipulation detection (Stage-1) and the fine manipulation segmentation (Stage-2). Initially, the LMFE module captures forgery clues of various content manipulation and creates a unified feature representation. Next, Stage-1, containing the attention region proposal network (RPN-A) and the prediction module, performs manipulation techniques classification and bounding box regression. In Stage-2, the skip structure fuses the multi-level information to enhance global feature representation. As prior knowledge, the bounding boxes from Stage- 1 guide Stage-2 to focus on local features for latter manipulation segmentation.
\subsection{Learnable manipulation feature extractor}
\begin{figure}
	\centering
	\includegraphics[width=0.9\linewidth]{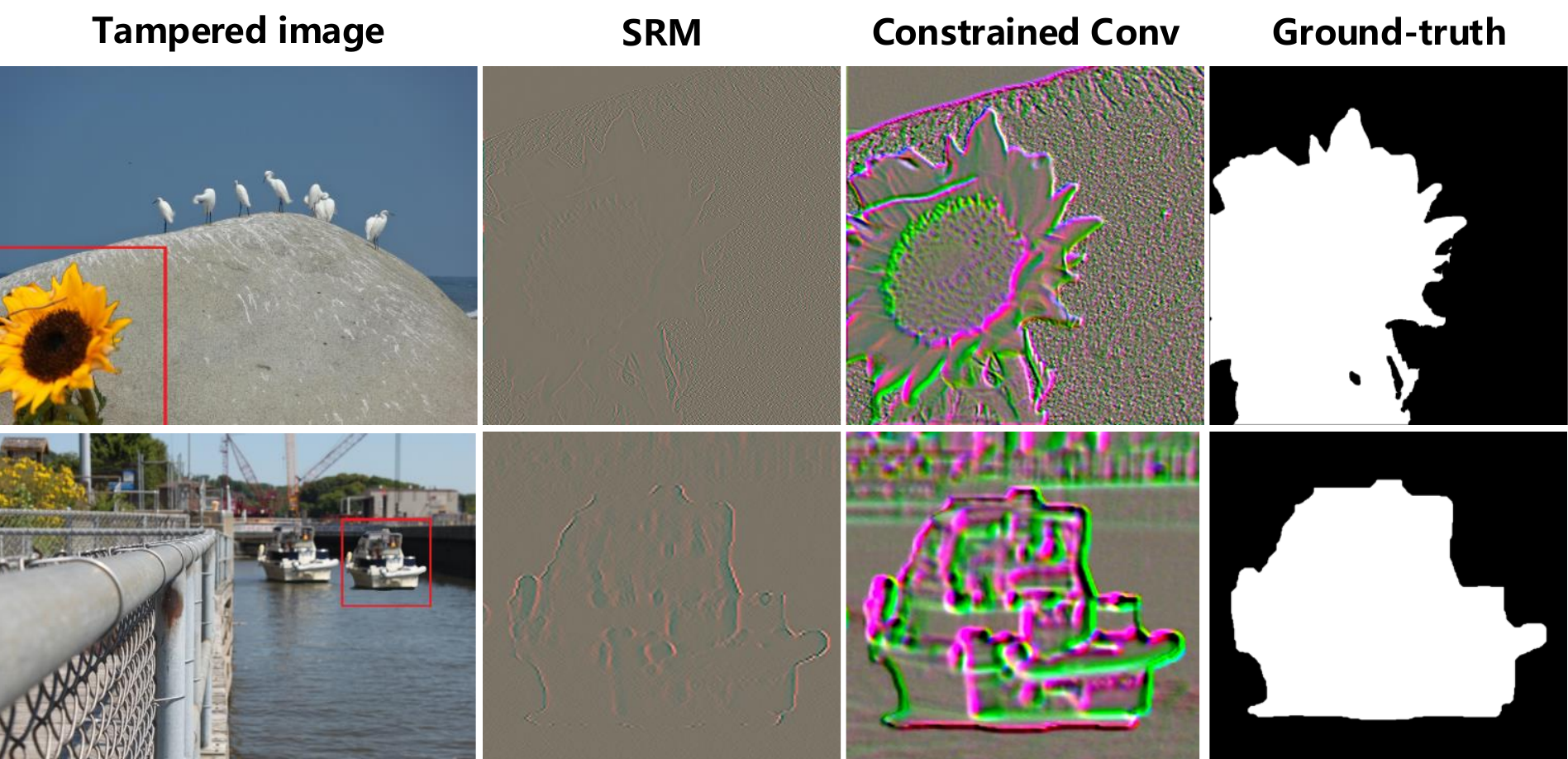}
	\vspace{-1em}
	\caption{Constrained Conv vs. SRM filters.}
	\label{fig:fig3}
	\vspace{-1.5em}
\end{figure}
The SRM filters \cite{DBLP:journals/tifs/FridrichK12} are widely utilized to capture manipulation clues (e.g., boundary anomaly, noise pattern inconsistencies). However, due to the limitation of handcrafted methods, the SRM filters are vulnerable to malicious attacks. In contrast, the constrained convolutional layer \cite{DBLP:journals/tifs/BayarS18} (Constrained Conv) adaptively learn manipulation features directly from data, which provides better universality and robustness. Specifically, the constraint is applied as follows:
\begin{equation}
\setlength\abovedisplayskip{4pt}
\setlength\belowdisplayskip{4pt}
\left\{
\begin{array}{lcl}
w_k(0, 0) = -1\\
\sum_{m, n\not=0}w_k(m, n)=1  
\end{array}  
\right.
\end{equation}
where $w_k$ denotes the $k^{th}$ convolution kernel, and (0, 0) is the center coordinate of $w_k$. $w_k$ are updated with the entire model, and then the above constraint process is performed.

As shown in Fig. \ref{fig:fig3}, we compare the visualization results of the SRM filters (same as \cite{DBLP:conf/cvpr/ZhouHMD18}) and the Constrained Conv (trained with our model). The second to fourth columns are the amplified regions of the red bounding boxes in the first column. On the whole, the Constrained Conv remains richer information than the SRM filters. We can clearly distinguish the object in the output of the Constrained Conv. From the details of the tampered regions, the Constrained Conv highlights the noise inconsistency between the authentic regions and the tampered regions (see Fig. \ref{fig:fig3} first-row third-column). It demonstrates that Constrained Conv feature has the similar effect of both RGB spatial feature and SRM feature.

According to the above comparative analysis, we design a learnable manipulation feature extractor (LMFE) based on the Constrained Conv to adaptively learn manipulation detection features. As shown in Fig. \ref{fig:fig2},  the LMFE takes a manipulated image as input, which is forward through the Constrained Conv layer for capturing rich manipulation clues. And then the Constrained Conv feature is input to the ResNet-101 for creating a unified feature representation of various content manipulation. Compared with \cite{DBLP:conf/cvpr/ZhouHMD18},the LMFE only has half number of parameters due to the single-stream design.
\subsection{Coarse Manipulation Detection (Stage-1)}
In Stage-1, the proposed model performs manipulation techniques classification and coarse manipulated region localization, which mainly contains the attention regional proposal network (RPN-A) and the prediction module.

\textbf{RPN-A.} As shown in Fig. \ref{fig:fig3}, although the Constrained Conv captures rich manipulation clues, it loses a lot of content information (e.g., color, brightness). The lack of content information weakens the inter-class discrimination of global features and ultimately impairs manipulation segmentation. 

To address this issue, we design an attention regional proposal network (RPN-A) to make the network learn feature representation with strong inter-class distinctive abilities. Specifically, a convolutional block attention module \cite{DBLP:conf/eccv/WooPLK18} (CBAM) is added to region proposal network \cite{DBLP:conf/nips/RenHGS15} (RPN). The CBAM first infers two attention maps along two separate dimensions, spatial and channel. Then, these two attention maps are multiplied to the input feature map for a more discriminative feature. Next, RPN-A utilizes the feature map from CBAM to propose the regions of interest (RoIs) that is the potential manipulated regions, where the tampered region is defined as foreground rather than the objects region \cite{DBLP:conf/cvpr/ZhouHMD18}.

For training RPN-A, the loss function is designed following \cite{DBLP:conf/nips/RenHGS15}. Formally, we define the loss of RPN-A as:\vspace{-0.5em}
\begin{eqnarray}
\mathcal{L}_{RPN-A}(\{p_i\}, \{t_i\})=\frac{1}{N_{cls}}\sum_{i}\mathcal{L}_{cls}(p_i,p_i^*)\nonumber\ \\	+\lambda\frac{1}{N_{reg}}\sum_{i}p_i^*\mathcal{L}_{reg}(t_i,t_i^*)
\vspace{-0.5em}
\end{eqnarray}
where $p_i$ and $p_i^*$ denote the classification probability and the label of anchor i. $t_i$ and $t_i^*$ denote the four dimensional coordinate of anchor i and ground-truth, respectively. $N_{cls}$ and $N_{reg}$ are the size of mini-batch and the number of anchors locations. $\mathcal{L}_{cls}$ is the cross-entropy classification loss. $\mathcal{L}_{reg}$ denotes smooth $L_1$ loss for the bounding boxes of proposals. $\lambda$ is a balancing parameter and defaults to 10.

\textbf{Prediction Module.} After RPN-A, the RoI Align \cite{DBLP:conf/iccv/HeGDG17} crops and resizes the local feature in RoI to the same size. And then, the fully connected and softmax layers are exploited to the manipulation techniques classification and the bounding box regression (coarse manipulation localization).

We use cross-entropy loss for final manipulation classification and smooth $L_1$ loss for final bounding box regression. The loss function of Stage-1 is defined as:\vspace{-0.5em}
\begin{eqnarray}
\mathcal{L}_{Stage-1}=\mathcal{L}_{RPN-A}+\mathcal{L}_{cls\_pred}+\mathcal{L}_{bbox\_pred}
\vspace{-0.5em}
\label{eq3}
\vspace{-0.5em}
\end{eqnarray}
where $\mathcal{L}_{RPN-A}$ is the loss of RPN-A, $\mathcal{L}_{cls\_pred}$ is the final classification loss, $\mathcal{L}_{bbox\_pred}$ is the final bounding box regression loss. Finally, the three losses care summed together to produce the loss function for Stage-1, as same as \cite{DBLP:conf/nips/RenHGS15}.
\subsection{Fine Manipulation Detection (Stage-2)}
In Stage-2, the coarse localization information from Stage-1 guides the model to further refine local features and performs manipulation segmentation on pixel level.

Considering there exists rich forgery clues on the boundary of tampered regions, we design a skip structure to introduce the low-level features ($Conv\_3x$) that contain more details and fuse it with high-level features (CBAM) by element-wise addition (see Fig. \ref{fig:fig2}). For matching the input size of the $Conv\_5x$ block, a convolutional layer with kernel size $1\times1$ expands the channel of the global enhanced features to 1024.

Meanwhile, Stage-2 further refines the local feature by exploiting the bounding box from Stage-1. Assume that the batch size of Stage-2 is $N$, and the size of bounding boxes is $N\times4$. First, the RoI Align crops and resizes the local enhanced feature to $N\times7\times7\times1024$. Next, local feature is input to the $Conv5\_x$ block of ResNet-101 for finer local features with size $N\times7\times7\times2048$. Finally, the decoder learns the mapping from local feature to pixel-wise prediction for tampered region segmentation. More specifically, a deconvolutional layer is exploited to upsample and decrease the channels of local feature (output size $N\times14\times14\times256$). As a transition, a convolutional layer with kernel size $1\times1$ further decreases the number of channels to 64. Since the manipulation classification has been performed in Stage-1, the class-agnostic segmentation is exploited for reducing the complexity of the model. Therefore, the output size of the second convolutional layer is $N\times14\times14\times2$. Finally, a softmax layer is exploited to predict the pixel-wise classification.

During training, we crop the tampered region on the ground truth mask and resize it to $14\times14$. Then, the average binary cross-entropy loss function is used to compute the loss between the predicted mask and the adjusted ground-truth.

Finally, we obtain the predicted content manipulation class and the binary mask of tampered regions. The total loss is defined as $\mathcal{L}_{total}=\mathcal{L}_{Stage-1}+\mathcal{L}_{Stage-2}$, where $\mathcal{L}_{total}$ denotes the total loss of Constrained R-CNN, $\mathcal{L}_{Stage-1}$ is the loss of Stage-1 defined in Eq. \ref{eq3}, and $\mathcal{L}_{Stage-2}$ is the average binary cross-entropy loss of Stage-2.

\section{Experiments}
\begin{table}[t]
	\small
	\begin{center}
		\caption{AP comparison on COCO synthetic dataset.} \label{tab:cap1}
		\vspace{0.5em}
		\begin{tabular}{l c}
			\hline
			AP & COCO Synthetic test\\
			\hline
			RGB-N \cite{DBLP:conf/cvpr/ZhouHMD18} & 0.745  \\
			Conv-C Net & 0.790  \\
			Conv-C Net + CBAM-34R & 0.779 \\
			Conv-C Net + CBAM-4 & 0.795 \\
			Conv-C Net + CBAM-4R & 0.801 \\
			\textbf{Conv-C Net + CBAM-R} & \textbf{0.806}\\
			\hline\hline
		\end{tabular}
	\end{center}
	\vspace{-2em}
\end{table}
In this section, we investigate the effectiveness of each part in Constrained R-CNN. We evaluate the proposed model on four image forensics benchmarks and compare the results with state-of-the-art methods. We also compare the influence of different data augmentation methods for our model.
\subsection{Implementation Details}
The proposed network is trained end-to-end. During training, we find the gradient will explode if the entire model is trained end-to-end directly. To overcome this issue, we first pre-train LMEF and Stage-1, where the ResNet-101 is initialized by ImageNet weights. On image forensics benchmarks, we train the entire model end-to-end with pre-trained weights. We will release the source code of our model$\footnote{https://github.com/HuizhouLi/Constrained-R-CNN}$.

The input image is resized so that the shorter side of the image is 600 pixels. Image flipping is used for data augmentation. The batch size of RPN-A proposal is 256 for training and 300 for testing. The batch size of Stage-2 is 4 for training and 8 for testing. We pre-train our model for 110K steps. Learning rate is initially set to $1e-3$, and then is reduced to $1e-4$ and $1e-5$ after 40K steps and 90K steps, respectively. All experiments are performed on single NVIDIA 1080Ti GPU.
\subsection{Pre-trained Model}
As current standard datasets do not have enough data for deep neural network training, some works utilize semantic segmentation datasets to create the manipulated image dataset \cite{DBLP:conf/cvpr/ZhouHMD18,DBLP:journals/tip/BappySNMR19}. For fair comparison, we use the same COCO synthetic dataset$\footnote{https://github.com/pengzhou1108/RGB-N}$ as \cite{DBLP:conf/cvpr/ZhouHMD18} to pre-train Stage-1. The output of the pre-trained model is bounding boxes with confidence scores denoting the probability of the box containing tampered regions. Following \cite{DBLP:conf/cvpr/ZhouHMD18}, we also use average precision (AP) for Stage-1 evaluation, same as the COCO detection evaluation.

We compare the AP of RGB-N and various model architectures in Table \ref{tab:cap1}. We reproduce the experiment of RGB-N on the same COCO synthetic dataset, the result is better than literature. The Conv-C Net is a single-stream Faster R-CNN model with the Constrained Conv. The string after 'CBAM' denotes the location of the CBAM module in Conv-C Net, where '3', '4' denote $Conv\_3x$ and $Conv\_4x$ respectively, the letter 'R' denotes the first convolutional layer in RPN.

The experimental results demonstrate that Constrained Conv significantly improves the AP with only half parameters. In addition, we explore the effect of the CBAM for our model. As same as the discussion in Section 4.3, adding a CBAM in RPN makes it learn a feature with strong inter-class distinctive ability. Compared with RGB-N, the Conv-C Net + CBAM-R (pre-trained model) improved the AP by 6.1\%.
\begin{table}[t]
	\small
	\begin{center}
		\caption{$F_1$ score comparison on four benchmarks.} \label{tab:cap2}	
		\vspace{0.5em}
		\begin{tabular}{l c c c c}
			\hline
			Methods&NIST16&COVER&Columbia&CASIA\\
			\hline
			NOI1 \cite{DBLP:journals/ivc/MahdianS09}&0.285&0.269&0.574&0.263   \\
			ELA \cite{ELA} & 0.236&0.222&0.470&0.214  \\
			MFCN \cite{DBLP:journals/jvcir/SalloumRK18} & 0.570&-&0.612&\textbf{0.541} \\
			RGB-N \cite{DBLP:conf/cvpr/ZhouHMD18} & 0.722&0.437&0.697&0.408 \\
			\hline
			Ours-Base & 0.920&0.750&0.709&0.469\\
			\textbf{Ours} &\textbf{0.927}&\textbf{0.757}&\textbf{0.790}&0.475\\
			\hline\hline
		\end{tabular}
	\end{center}
	\vspace{-1.8em}
\end{table}
\begin{table}[t]
	\small
	\begin{center}
		\caption{AUC comparison on four benchmarks.} \label{tab:cap3}
		\vspace{0.5em}
		\begin{tabular}{l c c c c}
			\hline
			Methods&NIST16&COVER&Columbia&CASIA\\
			\hline
			NOI1 \cite{DBLP:journals/ivc/MahdianS09}&0.487&0.587&0.545&0.612   \\
			ELA \cite{ELA} & 0.429&0.583&0.581&0.613  \\
			J-LSTM \cite{DBLP:conf/iccv/BappyRBNM17} & 0.764&0.614&-&- \\
			H-LSTM \cite{DBLP:journals/tip/BappySNMR19} &0.794&0.712&-&-\\
			RGB-N \cite{DBLP:conf/cvpr/ZhouHMD18} & 0.937&0.817&0.858&0.795\\
			MT-Net \cite{DBLP:conf/cvpr/0001AN19} &0.795&0.819&0.824&\textbf{0.817}\\
			\hline
			Ours-Base & 0.991&0.918&0.818&0.786\\
			\textbf{Ours} &\textbf{0.992}&\textbf{0.939}&\textbf{0.861}&0.789\\
			\hline\hline
		\end{tabular}
	\end{center}
	\vspace{-1em}
\end{table}
\subsection{Experiments on Standard Datasets}
We compare our method with different state-of-the-art methods on four manipulated image benchmarks: NIST16 \cite{NIST2016}, CASIA \cite{CASIA2.0}, COVERAGE \cite{7532339}, and Columbia datasets \cite{columbia}. The training and testing protocol of datasets is the same as \cite{DBLP:conf/cvpr/ZhouHMD18,DBLP:journals/jvcir/SalloumRK18,DBLP:conf/iccv/BappyRBNM17}. The Colombia dataset is not exploited for training, only for testing the model trained on the CASIA weights. 

Following conventional settings\cite{DBLP:journals/jvcir/SalloumRK18,DBLP:journals/mta/ZampoglouPK17}, we report the pixel-level $F_1$ score and the area under the ROC curve (AUC) for performance evaluation. For each output map, we vary different thresholds to get the maximum $F_1$ score as the final score, which follows the same protocol in \cite{DBLP:conf/cvpr/ZhouHMD18,DBLP:journals/jvcir/SalloumRK18,DBLP:journals/mta/ZampoglouPK17}

Table \ref{tab:cap2} describes the comparison of $F_1$ score between our method and the baselines. Table \ref{tab:cap3} shows the comparison of AUC. '-' denotes that the corresponding results are not provided in the literature. We obtain the results of NOI1 and ELA from \cite{DBLP:conf/cvpr/ZhouHMD18} and \cite{DBLP:journals/jvcir/SalloumRK18}. The results of MFCN, J-LSTM, H-LSTM, RGB-N, MT-Net are replicated from the original papers.

From the results in Table \ref{tab:cap2} and Table \ref{tab:cap3}, we can find that Constrained R-CNN exhibits significant improvement compared with state-of-the-art methods. Especially, the $F_1$ score of our model is increased by 0.205, 0.32, and 0.093 on the NIST16, COVERAGE, and Columbia datasets, with the growth rates of 28.4\%, 73.2\%, and 13.3\%, respectively. 

\begin{table}[t]
	\small
	\begin{center}
		\caption{Comparison of data augmentation methods. Flipping: Image flipping horizontally. Noise: Gaussian noise with the mean of 0 and the variance of 5. JPEG: JPEG compression with quality 70. Each entry is $F_1$/AUC score.} 
		\vspace{0.5em}
		\label{tab:cap4}
		\begin{tabular}{l c c c}
			\hline
			$F_1$/AUC&NIST16&COVER&CASIA \\
			\hline
			None & 0.917/0.989&0.742/0.918&0.467/0.768 \\
			Flipping &\textbf{0.927/0.992}&\textbf{0.757/0.939}&0.475/0.789 \\
			Flipping + Noise&0.926/0.991&0.736/0.933&0.445/0.753\\
			Flipping + JPEG&0.926/0.993&0.738/0.903&\textbf{0.498/0.805}\\
			All&0.923/0.991&0.646/0.884&0.485/0.784\\
			\hline\hline
		\end{tabular}
	\end{center}
	\vspace{-2em}
\end{table}
\begin{table}
	\begin{center}
		\caption{AP comparison on multi-class on NIST16 dataset.} 
		\vspace{0.5em}
		\label{tab:cap5}
		\begin{tabular}{l c c c c}
			\hline
			AP&Splicing&Copy-move&Removal&Mean \\
			\hline
			RGB-N \cite{DBLP:conf/cvpr/ZhouHMD18} & \textbf{0.960}&0.903&\textbf{0.939}&0.934  \\
			\textbf{Ours} &0.943& \textbf{0.999}&0.875&\textbf{0.939}\\
			\hline\hline
		\end{tabular}
	\end{center}
	\vspace{-1.5em}
\end{table}
In addition, we also demonstrate the effectiveness of the skip structure in Table \ref{tab:cap2} and Table \ref{tab:cap3}. Ours-Base is another architecture that is the same as Constrained R-CNN but without skip structure. Compared with Ours-Base, Constrained R-CNN significantly improves the performance of detection.

As shown in Table \ref{tab:cap4}, we compare the effects of different data augmentation method. Compared with other methods, image flipping provides significant improvement on two datasets. JPEG compression improves the performance of the model on CASIA dataset, and the AUC of Constrained R-CNN is comparable to state-of-the-art model \cite{DBLP:conf/cvpr/0001AN19}.
\subsection{Manipulation Techniques Classification Comparison}
We argue that image manipulation detection consists of classification and localization. The NIST16 dataset provides labels for three content manipulation techniques: splicing, copy-move, and removal. For fair comparison, we use the classification result on bounding box level.

As shown in Table \ref{tab:cap5}, Constrained R-CNN achieves state-of-the-art performance on the entire dataset. Compared with \cite{DBLP:conf/cvpr/ZhouHMD18}, Constrained R-CNN demonstrates significant improvements in copy-move detection and shows comparable performance to RGB-N on splicing detection.

\begin{figure}
	\centering
	\includegraphics[width=1\linewidth]{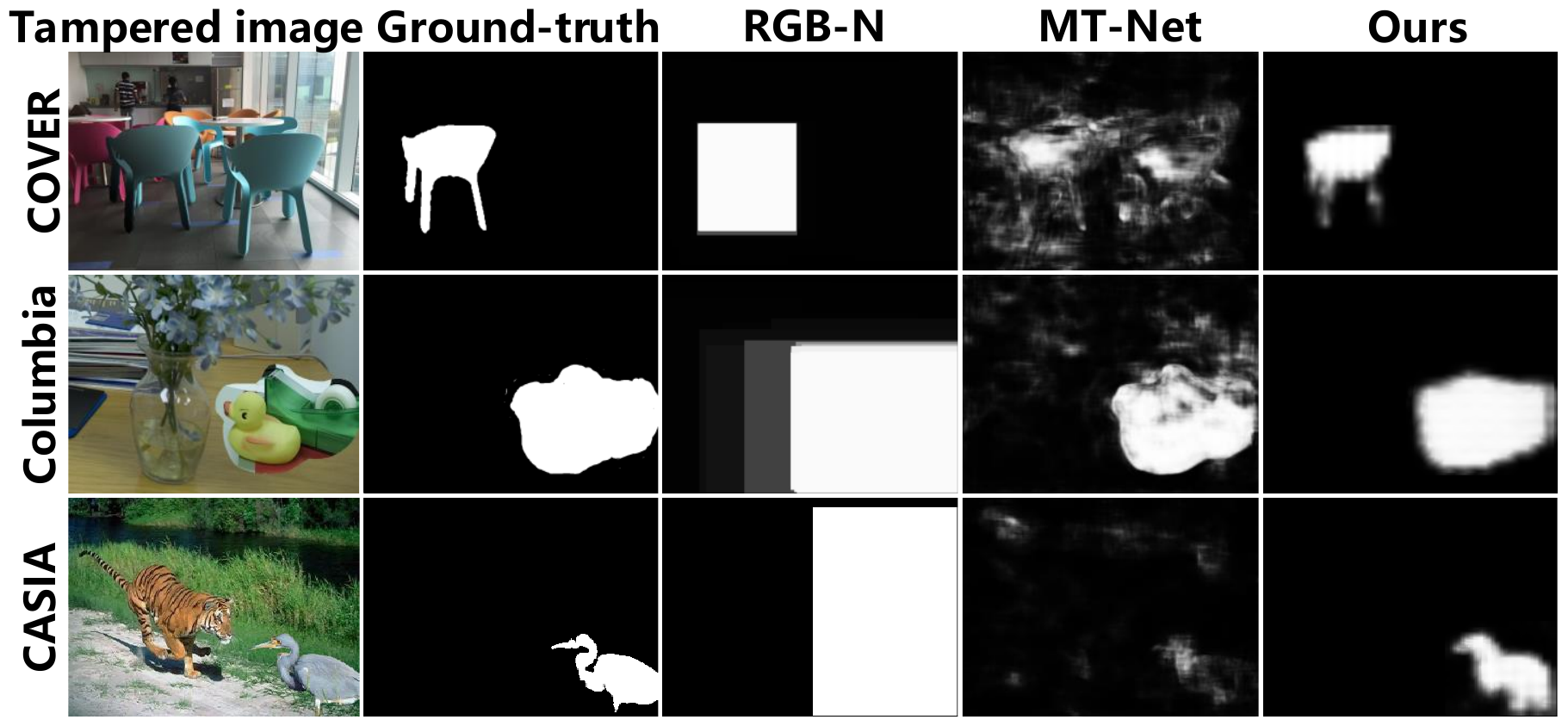}
	\vspace{-2.5em}
	\caption{Qualitative visualization of localization results.}
	\vspace{-1em}
	\label{fig:fig4}
\end{figure}
\begin{figure*}
	\centering
	\includegraphics[width=1\linewidth]{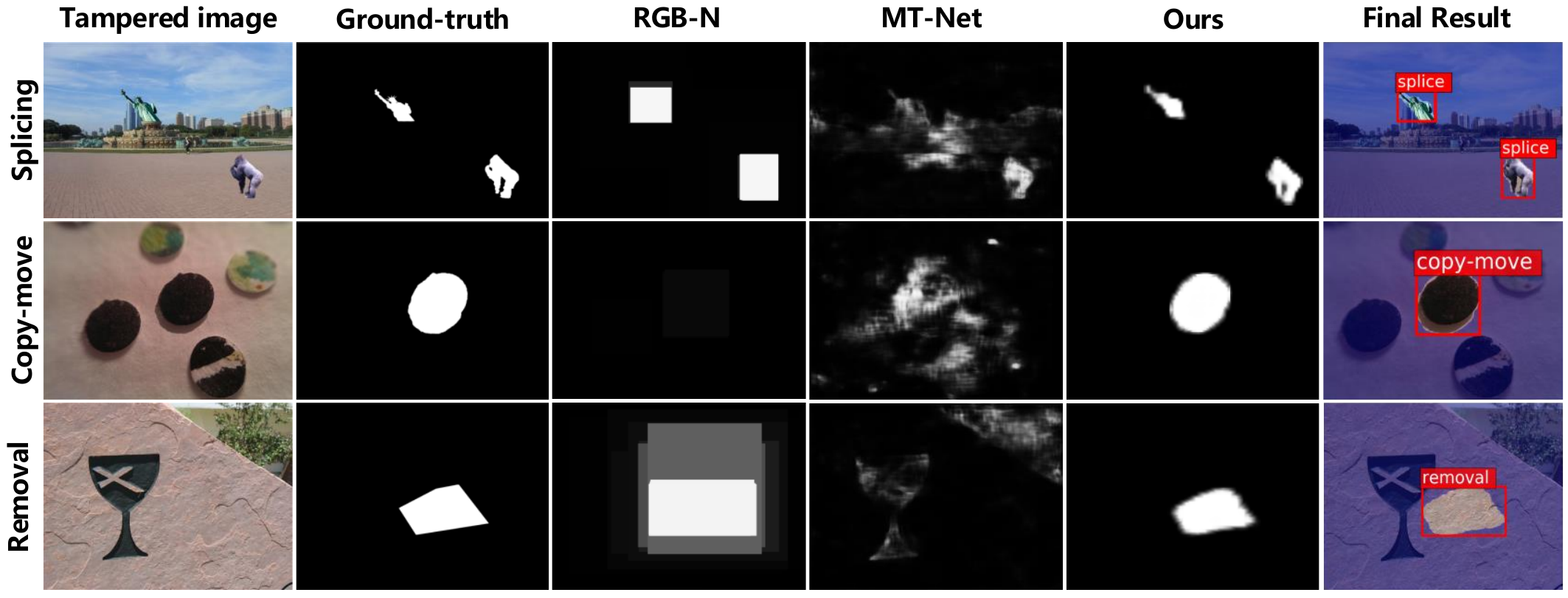}
	\vspace{-2.5em}
	\caption{Qualitative results for multi-class image manipulation detection on NIST16 dataset.}
	\label{fig:fig5}
	\vspace{-0.5em}
\end{figure*} 
\subsection{Qualitative Comparison}
In Fig. \ref{fig:fig4} and Fig. \ref{fig:fig5}, we compare the detection results wtih RGB-N \cite{DBLP:conf/cvpr/ZhouHMD18} and MT-Net \cite{DBLP:conf/cvpr/0001AN19}. Sincec there do not provide the label of manipulation technique in the COVERAGE, Colombia, and CASIA dataset, Fig. \ref{fig:fig4} only shows the manipulation localization results of them. In Fig. \ref{fig:fig5}, we compare the detection results of different manipulation techniques on NIST16 dataset. We also show the complete detection results of our model (the last column in Fig. \ref{fig:fig5}), which contain manipulation techniques classification and manipulation localization.

As shown in Fig. \ref{fig:fig4} and Fig. \ref{fig:fig5}, the constrained R-CNN produces good performance of manipulation techniques classification and better performance of manipulation localization than other models. This is because the coarse-to-fine design allows our model to perform detection tasks in the local regions, which avoids excessive noise in the output mask and ensures more accurate segmentation results.
%

\section{conclusions}
In this paper, we propose a coarse-to-fine architecture named Constrained R-CNN for image manipulation detection, which can simultaneously perform manipulation techniques classification and manipulation segmentation. Compared with previous methods, our model can capture manipulation clues directly from data without any handcrafted component, and thus is more general and effective for complex image forensics. Experiments on four manipulated image benchmarks demonstrate that our method achieves state-of-the-art performance in both manipulation classification and localization.

\section{acknowledgments}
This work was supported in part by National Natural Science Foundation
of China under Grant No. 61702176.
\bibliographystyle{IEEEbib}
\bibliography{Reference}

\end{document}